\documentclass[10pt,twocolumn,letterpaper]{article}

\usepackage{wacv}
\usepackage{times}
\usepackage{epsfig}
\usepackage{graphicx}
\usepackage{amsmath}
\usepackage{amssymb}
\usepackage{adjustbox}
\usepackage{booktabs}
\usepackage{multirow}
\usepackage[bookmarks=false]{hyperref}
\usepackage{caption}
\usepackage[algo2e]{algorithm2e}
\usepackage{algorithm}
\usepackage[noend]{algpseudocode}
\usepackage{multirow}
\usepackage{float}

\makeatletter
\newcommand{\appendixhead}%
{\centering {\huge Supplementary}%

\vspace{0.25in}
\centering
\scalebox{.8}{
    \includegraphics[width=2.7in,height=2.88in,keepaspectratio]{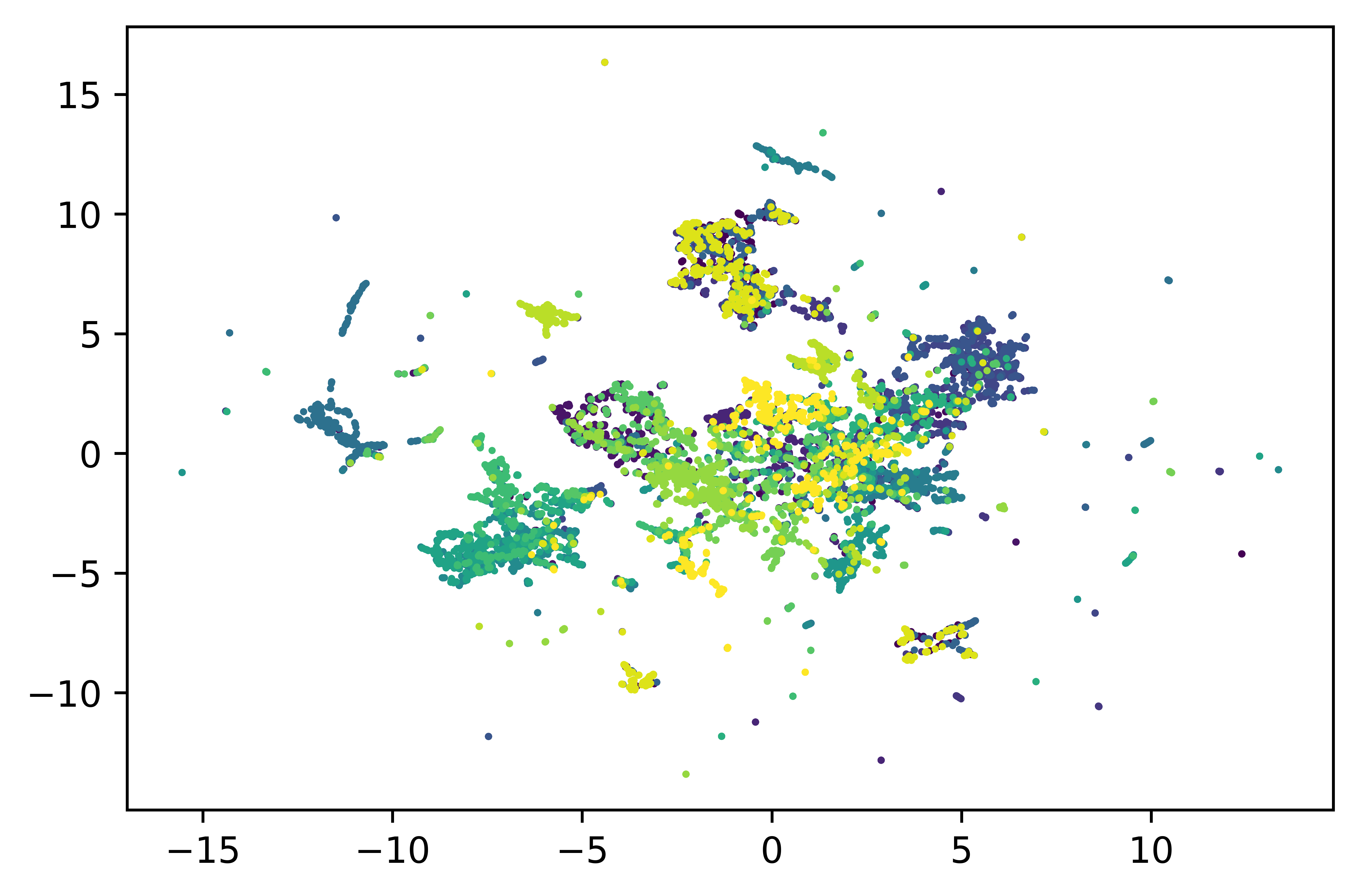} 
    \includegraphics[width=2.7in,height=2.88in,keepaspectratio]{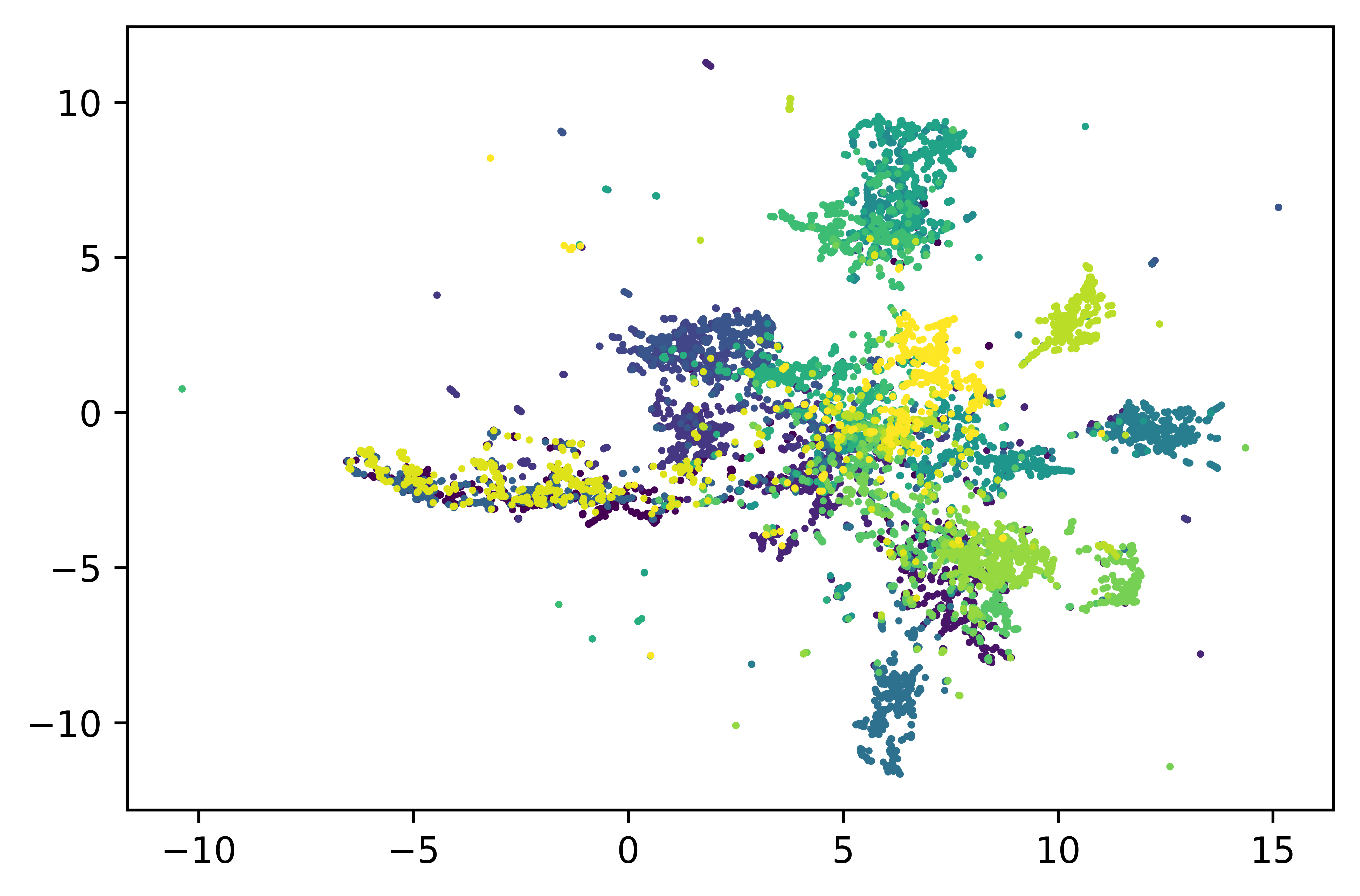} 
    \includegraphics[width=2.7in,height=2.88in,keepaspectratio]{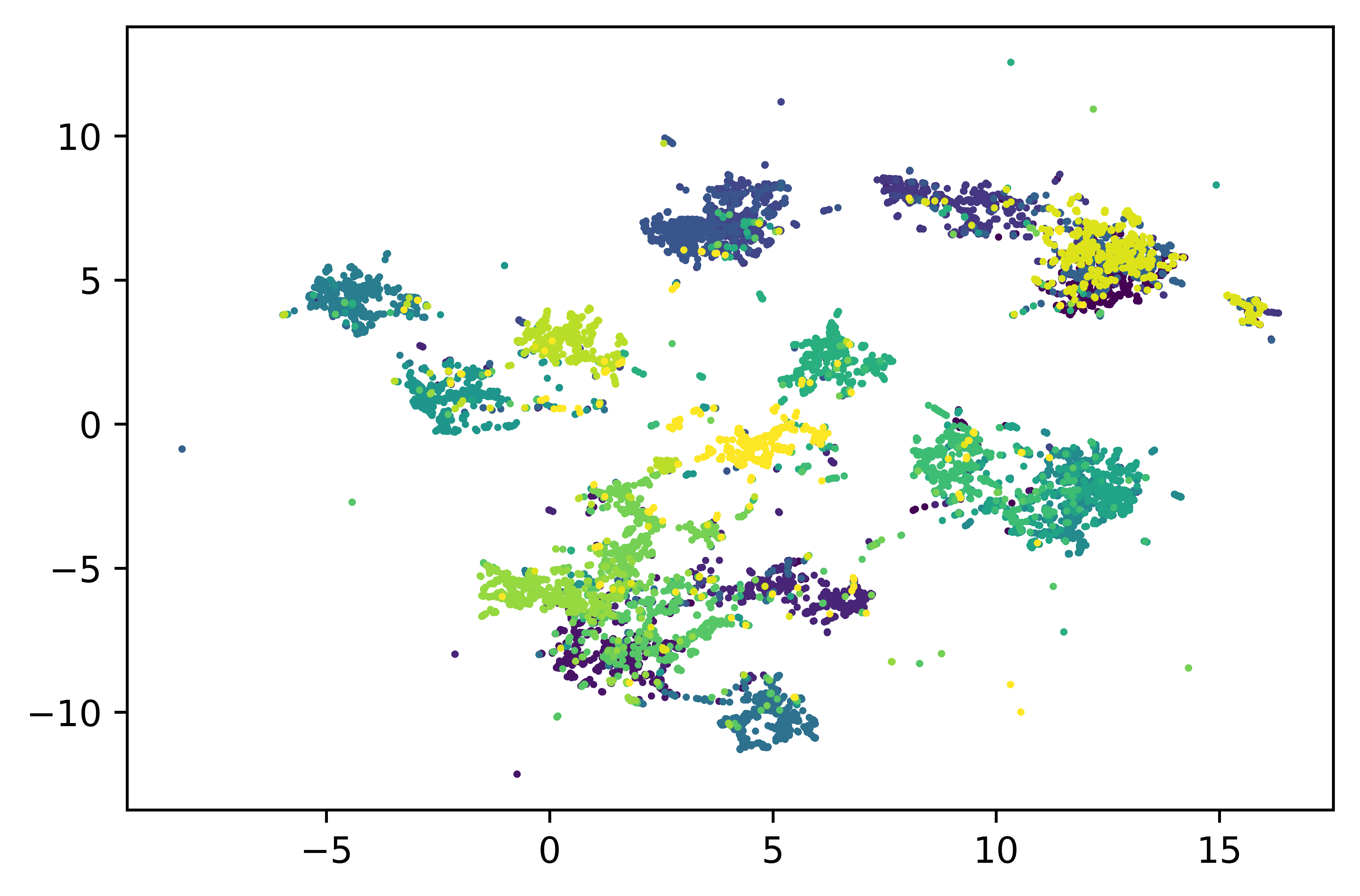}
}
    \captionof{figure}{\footnotesize{UMAP (2-dim) \cite{umaplearn} plot for feature vectors of examples from novel classes of CIFAR-FS using Baseline++ \cite{chen2019closerfewshot}, Rotation, \textit{S2M2$_R$} (top to bottom).}}
    \label{cifar2-dplot}

\begin{center}
\scalebox{1.0}{
\begin{tabular}{@{}|c|l|l|l|l|l|l|l|l|@{}}
\toprule
\multirow{2}{*}{Method} & \multicolumn{2}{c|}{5-way}  & \multicolumn{2}{c|}{10-way}    & \multicolumn{2}{c|}{15-way}     & \multicolumn{2}{c|}{20-way}    \\
                        & 1-shot        & 5-shot         & 1-shot         & 5-shot 
                        & 1-shot         & 5-shot
                        & 1-shot        & 5-shot         \\ \midrule
Baseline++ & 67.5 & 80.08 & 53.39 & 68.89 & 44.73 & 60.59 & 38.22  & 54.68   \\ \midrule
Manifold Mixup    & 69.45 & 83.31      & 57.06         & 75.53          &  49.04        &  68.60           & 43.54         & 62.80         \\
Rotation      & 70.5 & 84.03          & 57.37         & 73.60           & 48.49           & 66.25          & 42.28         & 61.10         \\
\textit{S2M2$_R$}        & \textbf{74.45} & \textbf{87.50}         & \textbf{62.28} & \textbf{78.47} & \textbf{53.49} & \textbf{71.88} & \textbf{47.59} & \textbf{66.37} \\ \bottomrule
\end{tabular}
}
\end{center}
\captionof{table}{\footnotesize{Mean few-shot accuracy on CIFAR-FS as $N$ increases in $N$-way $K$-shot classification.}}
\label{table:N_Waycifar}

\bigskip}
\makeatother

\wacvfinalcopy 

\ifwacvfinal\fi
\begin{document}

\title{Charting the Right Manifold: Manifold Mixup for Few-shot Learning
}

\author{Puneet Mangla\thanks{Authors contributed equally   }   \thanks{Work done during Adobe MDSR internship }  $^{2}$ \\
{\tt\small cs17btech11029@iith.ac.in}
\and
Mayank Singh$^{*1}$ \\
{\tt\small msingh@adobe.com}
\and
Abhishek Sinha$^{*1}$ \\
{\tt\small abhsinha@adobe.com}
\and
Nupur Kumari$^{*1}$ \\
{\tt\small nupkumar@adobe.com}
\and
Vineeth N Balasubramanian$^{2}$ \\
{\tt\small vineethnb@iith.ac.in}
\and
Balaji Krishnamurthy$^{1}$ \\
{\tt\small kbalaji@adobe.com}\\
\and
1. Media and Data Science Research lab, Adobe\\
2. IIT Hyderabad, India
}

\maketitle
\ifwacvfinal\thispagestyle{empty}\fi

\begin{abstract}

Few-shot learning algorithms aim to learn model parameters capable of adapting to unseen classes with the help of only a few labeled examples. A recent regularization technique - Manifold Mixup focuses on learning a general-purpose representation, robust to small changes in the data distribution. Since the goal of few-shot learning is closely linked to robust representation learning, we study Manifold Mixup in this problem setting. Self-supervised learning is another technique that learns semantically meaningful features, using only the inherent structure of the data. This work investigates the role of learning relevant feature manifold for few-shot tasks using self-supervision and regularization techniques. We observe that regularizing the feature manifold, enriched via self-supervised techniques, with Manifold Mixup significantly improves few-shot learning performance.  
We show that our proposed method \textit{S2M2} beats the current state-of-the-art accuracy on standard few-shot learning datasets like CIFAR-FS, CUB, \textit{mini}-ImageNet and {\textit{tiered}-ImageNet} by $3-8\%$. Through extensive experimentation, we show that the features learned using our approach generalize to complex few-shot evaluation tasks, cross-domain scenarios and are robust against slight changes to data distribution.
\end{abstract}

\section{Introduction}
Deep convolutional networks (CNN's) have become a regular ingredient for numerous contemporary computer vision tasks. They have been applied to tasks such as object recognition, semantic segmentation, object detection \cite{introconv1,introconv2,maskrnn,he2016deep,krizhevsky2012ImageNet} to achieve state-of-the-art performance.
However, the at par performance of deep neural networks requires huge amount of supervisory examples for training. Generally, labeled data is scarcely available and data collection is expensive for several problem statements. 
Hence, a major research effort is being dedicated to fields such as transfer learning, domain adaptation, semi-supervised and unsupervised learning \cite{domain_unsupervised_2015,domain_cluster_2010,transfer2010} to alleviate this requirement of enormous amount of examples for training.

A related problem which operates in the low data regime is few-shot classification. In few-shot classification, the model is trained on a set of classes (base classes) with abundant examples in a fashion that promotes the model to classify unseen classes (novel classes) using few labeled instances. The motivation for this stems from the hypothesis that an appropriate prior should enable the learning algorithm to solve consequent tasks more easily. Biologically speaking, humans have a high capacity to generalize and extend the prior knowledge to solve new tasks using only small amount of supervision. 
One of the promising approach to few-shot learning utilizes meta-learning framework to optimize for such an initialization of model parameters such that adaptation to the optimal weights of classifier for novel classes can be reached with few gradient updates \cite{ravi2016optimization, finn2017model, leo2019, nichol2018first}. Some of the work also includes leveraging the information of similarity between images \cite{vinyals2016matching,snell2017prototypical,sung2018learning,bertinetto2018meta,garcia2017few} and augmenting the training data by hallucinating additional examples \cite{hariharan2017low, wang2018low, fewshot2018hallo}.
Another class of algorithms \cite{fewshot2018cc,fewshot2018cc2} learns to directly predict the weights of the classifier for novel classes.

Few-shot learning methods are evaluated using $N$-way $K$-shot classification framework where $N$ classes are sampled from a set of novel classes (not seen during training) with $K$ examples for each class.
Usually, the few-shot classification algorithm has two separate learning phases. In the first phase, the training is performed on base classes to develop robust and general-purpose representation aimed to be useful for classifying novel classes. The second phase of training exploits the learning from previous phase in the form of a prior to perform classification over novel classes. The transfer learning approach serves as the baseline which involves training a classifier for base classes and then subsequently learning a linear classifier on the penultimate layer of the previous network to classify the novel classes \cite{chen2019closerfewshot}.


Learning feature representations that generalize to novel classes is an essential aspect of few-shot learning problem. This involves learning a feature manifold that is relevant for novel classes. Regularization techniques enables the models to generalize to unseen test data that is disjoint from training data. It is frequently used as a supplementary technique alongside standard learning algorithms \cite{bn_2015_generalize,drop_2012_generalize,bishop_1995_generalize,verma2019manifold,zhang2018mixup}.
In particular for classification problems, Manifold Mixup \cite{verma2019manifold} regularization leverages interpolations in deep hidden layer to improve hidden representations and decision boundaries at multiple layers.


In Manifold Mixup \cite{verma2019manifold}, the authors show improvement in classification task over standard image deformations and augmentations. Also, some work in self-supervision \cite{Spyros2018rotate,s4l2019,exemplar2014} explores to predict the type of augmentation applied and enforces feature representation to become invariant to image augmentations to learn robust visual features. Inspired by this link, we propose to unify the training of few-shot classification with self-supervision techniques and Manifold Mixup \cite{verma2019manifold}. The proposed technique employs self-supervision loss over the given labeled data unlike in semi-supervised setting that uses additional unlabeled data and hence our approach doesn't require any extra data for training.

Many of the recent advances in few-shot learning exploit the meta-learning framework, which simulates the training phase as that of the evaluation phase in the few-shot setting.
However, in a recent study \cite{chen2019closerfewshot}, it was shown that learning a cosine classifier on features extracted from deeper networks also performs quite well on few-shot tasks. Motivated by this observation, we focus on utilizing self-supervision techniques augmented with Manifold Mixup in the domain of few-shot tasks using cosine classifiers.


Our main contributions in this paper are the following: 
\begin{itemize}
    
    
    
    
    
    
    \item We find that the regularization technique of Manifold Mixup \cite{verma2019manifold} being robust to small changes in data distribution enhances the performance of few-shot tasks.
    
    \item We show that adding self-supervision loss to the training procedure, enables robust semantic feature learning that leads to a significant improvement in few-shot classification. We use rotation \cite{Spyros2018rotate} and exemplar \cite{exemplar2014} as the self-supervision tasks.

    \item We observe that applying Manifold Mixup regularization over the feature manifold enriched via the self-supervision tasks further improves the performance of few-shot tasks. The proposed methodology (\textit{S2M2}) outperforms the state-of-the-art methods by 3-8\% over the CIFAR-FS, CUB, \textit{mini}-ImageNet and {\textit{tiered}-ImageNet} datasets.
    
    \item We conduct extensive ablation studies to verify the efficacy of the proposed method. We find that the improvements made by our methodology become more pronounced with increasing $N$ in the $N$-way $K$-shot evaluation and also
    in the cross-domain evaluation.
    
\end{itemize}


\section{Related Work}
\label{related_work}

\begin{figure*}[t]
\centering
\scalebox{.95}{
    \includegraphics[width=2.6in,height=1.4in]{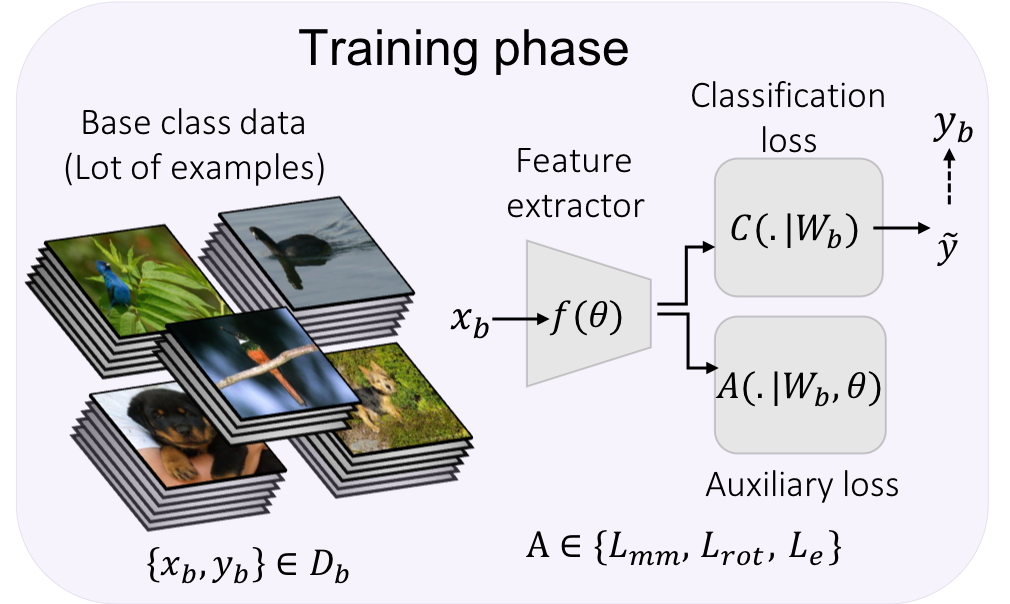} 
    \includegraphics[width=2.1in,height=1.4in]{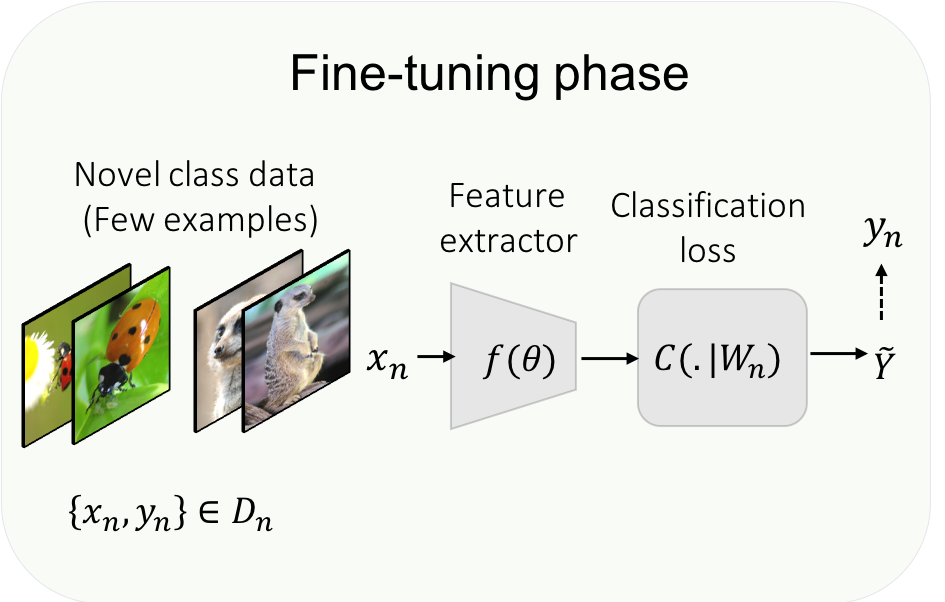} 
    \includegraphics[width=2.1in,height=1.4in]{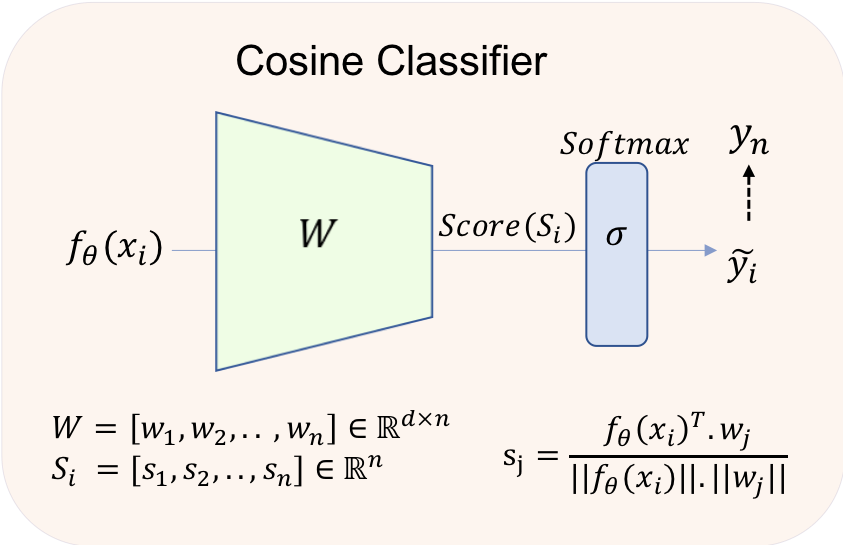} 
    }
    \captionof{figure}{\footnotesize{Flowchart for our proposed approach (\textit{S2M2}) for few-shot learning. The auxiliary loss is derived from Manifold Mixup regularization and self-supervision tasks of rotation and exemplar.}}
    \label{fig:self-supervised-few-shot}
\end{figure*}

Our work is associated with various recent development made in learning robust general-purpose visual representations, specifically few-shot learning, self-supervised learning and generalization boosting techniques. 

\paragraph{Few-shot learning:}
Few-shot learning involves building a model using available training data of base classes that can classify unseen novel classes using only few examples. Few-shot learning approaches can be broadly divided into three categories - gradient based methods, distance metric based methods and hallucination based methods.

Some gradient based methods \cite{ravi2016optimization, andrychowicz2016learning} aim to use gradient descent to quickly adapt the model parameters suitable for classifying the novel task. The initialization based methods \cite{finn2017model, leo2019, nichol2018first} specifically advocate to learn a suitable initialization of the model parameters, such that adapting from those parameters can be achieved in a few gradient steps.
Distance metric based methods leverage the information about similarity between images to classify novel classes with few examples.
The distance metric can either be cosine similarity \cite{vinyals2016matching}, euclidean distance \cite{snell2017prototypical}, CNN based distance module\cite{sung2018learning}, ridge regression\cite{bertinetto2018meta} or graph neural network\cite{garcia2017few}.
Hallucination based methods \cite{hariharan2017low, wang2018low, fewshot2018hallo} augment the limited training data for a new task by generating or hallucinating new data points.

Recently, \cite{chen2019closerfewshot} introduced a modification for the simple transfer learning approach, where they learn a cosine classifier \cite{fewshot2018cc,fewshot2018cc2} instead of a linear classifier on top of feature extraction layers. The authors show that this simple approach is competitive with several proposed few-shot learning approaches if a deep backbone network is used to extract the feature representation of input data.

\paragraph{Self-supervised learning:}
This is a general learning framework which aims to extract supervisory signals by defining surrogate tasks using only the structural information present in the data. In the context of images, a pretext task is designed such that optimizing it leads to more semantic image features that can be useful for other vision tasks. Self-supervision techniques have been successfully applied to diverse set of domains, ranging from robotics to computer vision \cite{robot2018,robot2018_2,TCN2017,crossmodal2018,multisensory2018}. In the context of visual data, the surrogate loss functions can be derived by leveraging the invariants in the structure of the image. In this paper, we focus on self-supervised learning techniques to enhance the representation and learn a relevant feature manifold for few-shot classification setting. We now briefly describe the recent developments in self-supervision techniques. 

C. Doersch \etal \cite{doersch2015seminal} took inspiration from spatial context of a image to derive supervisory signal by defining the surrogate task of relative position prediction of image patches. Motivated by the task of context prediction, the pretext task was extended to predict the permutation of the shuffled image patches \cite{Noroozi2018jigsaw,Nathan2018improve,Noroozi2018improve}. \cite{Spyros2018rotate} leveraged the rotation in-variance of images to create the surrogate task of predicting the rotation angle of the image. Also, the authors of \cite{rotate_decouple_2019} proposed to decouple representation learning of the rotation as pretext task from class discrimination to obtain better results. Along the lines of context-based prediction, \cite{painting2018} uses generation of the contents of image region based on context pixel (i.e. in-painting) and in \cite{color2018,color_2018_1} the authors propose to use gray-scale image colorization as a pretext task.

Apart from enforcing structural constraints, \cite{cluster2018self} uses cluster assignments as supervisory signals for unlabeled data and works by alternating between clustering of the image descriptors and updating the network by predicting the cluster assignments. \cite{moving_object2017} defines pretext task that uses low-level motion-based grouping cues to learn visual representation. Also, \cite{Noroozi2017count} proposes to obtain supervision signal by enforcing the additivity of visual primitives in the patches of images and \cite{oord2018cpc} proposed to learn feature representations by predicting the future in latent space by employing auto-regressive models.

Some of the pretext tasks also work by enforcing constraints on the representation of the feature. A prominent example is the exemplar loss from \cite{exemplar2014} that promotes representation of image to be invariant to image augmentations. Additionally, some research effort have also been put in to define the pretext task as a combination of multiple pretext task \cite{doersch2017multitask,damage_jigsaw_2018}. For instance, in \cite{damage_jigsaw_2018} representation learning is augmented with pretext tasks of jigsaw puzzle \cite{Noroozi2018jigsaw}, colorization \cite{color2018,color_2018_1} and in-painting \cite{painting2018}. 

\paragraph{Generalization:}
Employing regularization techniques for training deep neural networks to improve their generalization performances have become standard practice in the deep learning community. Few of the commonly used regularization techniques are - dropout \cite{srivastava2014dropout}, cutout \cite{devries2017improved}, Mixup \cite{zhang2018mixup}, Manifold Mixup \cite{verma2019manifold}. Mixup \cite{zhang2018mixup} is a specific case of Manifold Mixup \cite{verma2019manifold} where the interpolation of only input data is applied. The authors in \cite{verma2019manifold} claim that Manifold Mixup leads to smoother decision boundaries and flattens the class representations thereby leading to feature representation that improve the performance over a held-out validation dataset. We apply a few of these generalization techniques during the training of the backbone network over the base tasks and find that the features learned via such regularization lead to better generalization over novel tasks too. Authors of \cite{overview_2015_generalize} provide a summary of popular regularization techniques used in deep learning.
\vspace{-4pt}
\section{Methodology}\label{methodology}
\vspace{-3pt}
The few-shot learning setting is formalized by the availability of a dataset with data-label pairs $\mathcal{D} = \{(\textbf{x}_i,y_i): i=1,\cdots,m\}$ where $\textbf{x} \in \mathbb{R}^d$ and $y_i \in C$, $C$ being the set of all classes. We have sufficient number of labeled data in a subset of $C$ classes (called \textit{base classes}), while very few labeled data for the other classes in $C$ (called \textit{novel classes}).
Few-shot learning algorithms generally train in two phases: the first phase consists of training a network over base class data $\mathcal{D}_b=\{(\textbf{x}_i,y_i), i=1,\cdots,m_b\}$ where $\{y_i \in C_b \subset C\}$ to obtain a feature extractor, and the second phase consists of adapting the network for novel class data $\mathcal{D}_n=\{(\textbf{x}_i,y_i), i=1,\cdots,m_n\}$ where $\{y_i \in C_n \subset C\}$ and $C_b \cup C_n = C$. We assume that there are $N_b$ \textit{base} classes (cardinality of $C_b$) and $N_n$ \textit{novel} classes (cardinality of $C_n$). The general goal of few-shot learning algorithms is to learn rich feature representations from the abundant labeled data of base classes $N_b$, such that the features can be easily adapted for the novel classes using only few labeled instances.

In this work, in the first learning stage, we train a $N_{b}$-way neural network classifier:
\begin{equation}
g = c_{W_b} \circ f_{\theta} 
\label{eqn_nn_model}
\end{equation}
on $\mathcal{D}_b$, where $c_{W_b}$ is a cosine classifier \cite{fewshot2018cc,fewshot2018cc2} and $f_{\theta}$ is the convolutional feature extractor, with $\theta$ parametrizing the neural network model. The model is trained with classification loss and an additional auxiliary loss which we explain soon. The second phase involves fine-tuning of the backbone model, $f_{\theta}$, by freezing the feature extractor layers and training a new $N_n$-way cosine classifier $c_{W_n}$ on data from $k$ randomly sampled novel classes in $\mathcal{D}_n$ with only classification loss. Figure \ref{fig:self-supervised-few-shot} provides an overview of our approach \textit{S2M2} for few-shot learning .



Importantly, in our proposed methodology, we leverage self-supervision and regularization techniques \cite{verma2019manifold,Spyros2018rotate,exemplar2014} to learn general-purpose representation suitable for few-shot tasks. We hypothesize that using robust features which describes the feature manifold well is important to obtain better performance over the novel classes in the few-shot setting. In the subsequent subsections, we describe our training procedure to use self-supervision methods (such as rotation \cite{Spyros2018rotate} and exemplar \cite{exemplar2014}) to obtain a suitable feature manifold, following which using Manifold Mixup regularization \cite{verma2019manifold} provides a robust feature extractor backbone. We empirically show that this proposed methodology achieves the new state-of-the-art result on standard few-shot learning benchmark datasets.



\subsection{Manifold Mixup for Few-shot Learning}
Higher-layer representations in neural network classifiers have often been visualized as lying on a meaningful manifold, that provide the relevant geometry of data to solve a given task \cite{bengio2013representation}. Therefore, linear interpolation of feature vectors in that space should be relevant from the perspective of classification. With this intuition, Manifold Mixup \cite{verma2019manifold}, a recent work, leverages linear interpolations in neural network layers to help the trained model generalize better.
In particular, given input data $\textbf{x}$ and $\textbf{x}'$ with corresponding feature representations at layer $l$ given by $f_{\theta}^l(\textbf{x})$ and $f_{\theta}^l(\textbf{x}')$ respectively. Assuming we use Manifold Mixup on the base classes in our work, the loss for training $L_{mm}$ is then formulated as:
\begin{equation}
    L_{mm} = \mathbb{E}_{(x,y) \in \mathcal{D}_b} \Big[ L \big( Mix_{\lambda}(f_{\theta}^l(\textbf{x}), f_{\theta}^l(\textbf{x}')), Mix_{\lambda}(y, y')\big) \Big]
\label{eq:manifold-mixup}
\end{equation}
where 
\begin{equation}
    Mix_{\lambda}(a,b) = \lambda \cdot a + (1-\lambda) \cdot b 
\label{eq:eqn_mixup}
\end{equation}
The mixing coefficient $\lambda$ is sampled from a $\beta (\alpha , \alpha)$ distribution and loss $L$ is standard cross-entropy loss. We hypothesize that using Manifold Mixup on the base classes provides robust feature presentations that lead to state-of-the-art results in few-shot learning benchmarks.

Training using loss $L_{mm}$ encourages the model to predict less confidently on linear interpolations of hidden representations. This encourages the feature manifold to have broad regions of low-confidence predictions between different classes and thereby smoother decision boundaries, as shown in \cite{verma2019manifold}. Also, models trained using this regularizer lead to flattened hidden representations for each class with less number of directions of high variance i.e. the representations of data from each class lie in a lower dimension subspace. The above-mentioned characteristics of the method make it a suitable regularization technique for generalizing to tasks with potential distribution shifts.


\subsection{Charting the Right Manifold}
We observed that Manifold Mixup does result in higher accuracy on few-shot tasks, as shown in Section \ref{results}. However, it still lags behind existing state-of-the-art performance, which begs the question: \textit{Are we charting the right manifold?} In few-shot learning, novel classes introduced during test time can have a different data distribution when compared to base classes. In order to counter this distributional shift, we hypothesize that it is important to capture the right manifold when using Manifold Mixup for the base classes. To this end, we leverage self-supervision methods.
Self-supervision techniques have been employed recently in many domains for learning rich, generic and meaningful feature representations. We show that the simple idea of adding auxiliary loss terms from self-supervised techniques while training the base classes provides feature representations that significantly outperform state-of-the-art for classifying on the novel classes. We now describe the self-supervised methods used in this work.

\subsubsection{Self-Supervision: Towards the Right Manifold}
\label{selfsupervision}
In this work, we use two pretext tasks that have recently been widely used for self-supervision to support our claim. We describe each of these below.
\paragraph{Rotation \cite{Spyros2018rotate}:}
In this self-supervised task, the input image is rotated by different angles, and the auxiliary aim of the model is to predict the amount of rotation applied to image. In the image classification setting, an auxiliary loss (based on the predicted rotation angle) is added to the standard classification loss to learn general-purpose representations suitable for image understanding tasks. In this work, we use a $4$-way linear classifier, $c_{W_r}$, on the penultimate feature representation $f_{\theta}(\textbf{x}^r)$ where $\textbf{x}^r$ is the image $x$ rotated by $r$ degrees and $r \in C_R = \{0^{\circ},90^{\circ},180^{\circ},270^{\circ}\}$, to predict one of $4$ classes in $C_R$. In other words, similar to Eqn \ref{eqn_nn_model}, our pretext task model is given by $g_r = c_{W_r} \circ f_{\theta}$. The self-supervision loss is given by:
\begin{equation}
    L_{rot} = \frac{1}{|C_R|}*\sum\limits_{\textbf{x} \in \mathcal{D}_b} \sum\limits_{r \in C_R} L(c_{W_r}(f_{\theta}(\textbf{x}^r)) ,r)
\label{eqn_rot_selfsup_loss}
\end{equation}

\begin{equation}
    L_{class} = \mathbb{E}_{(x,y) \in \mathcal{D}_b , r \in C_R}  \big[ L (x^r,y) \big]
\label{eq:class-loss}
\end{equation}

\noindent where $|C_R|$ denotes the cardinality of $C_R$. As the self-supervision loss is defined over the given labeled data of $\mathcal{D}_b$, no additional data is required to implement this method. $L$ is the standard cross-entropy loss, as before.

\paragraph{Exemplar \cite{exemplar2014}:}
Exemplar training aims at making the feature representation invariant to a wide range of image transformations such as translation, scaling, rotation, contrast and color shifts. In a given mini-batch $M$, we create $4$ copies of each image through random augmentations. These $4$ copies are the positive examples for each image and every other image in the mini-batch is a negative example. We then use hard batch triplet loss  \cite{hardbatchtripletloss} with soft margin on $f_{\theta}(\textbf{x})$ on the mini-batch to bring the feature representation of positive examples close together. Specifically, the loss is given as:
\begin{equation}
\begin{aligned}
    L_{e} = & \frac{1}{4*|M|} \sum_{\textbf{x}\in M } \sum_{k=1}^{4} \log \bigg(1 + \exp\big(- \max_{p \in \{1,\cdots,4\}} D\big(\textbf{x}_k^i,\textbf{x}_p^i\big) \\
    & + \min_{p \in \{1..4\}, i \neq j}
        D(\textbf{x}_k^i,\textbf{x}_p^j)\big) \bigg)
\end{aligned}
\label{eqn_exemplar_selfsup_loss}
\end{equation}
Here, $D$ is the Euclidean distance in the feature representation space $f_\theta(\textbf{x})$ and $\textbf{x}_k^i$ is the $k^{th}$ exemplar of $\textbf{x}$ with class label $i$ (the appropriate augmentation). The first term inside the $\exp$ term is the maximum among distances between an image and its positive examples which we want to reduce. The second term is the minimum distance between the image and its negative examples which we want to maximize. 
\subsubsection{\textit{S2M2}: Self-Supervised Manifold Mixup}
\label{ourapproach}
The few-shot learning setting relies on learning robust and generalizable features that can separate base and novel classes. An important means to this end is the ability to compartmentalize the representations of base classes with generous decision boundaries, which allow the model to generalize to novel classes. Manifold Mixup provides an effective methodology to flatten representations of data from a given class into a compact region, thereby supporting this objective. However, while \cite{verma2019manifold} claims that Manifold Mixup can handle minor distribution shifts, the semantic difference between base and novel classes in the few-shot setting may be more than what it can handle. We hence propose the use of self-supervision as an auxiliary loss while training the base classes, which allows the learned backbone model, $f_{\theta}$, to provide feature representations with sufficient decision boundaries between classes, that allow the model to extend to the novel classes. This is evidenced in our results presented in Section \ref{results}. Our overall methodology is summarized in the steps below, and the pseudo-code of the proposed approach for training the backbone is presented in Algorithm \ref{feature_adv_train_algo}.
\paragraph{Step 1: Self-supervised training:} Train the backbone model using self-supervision as an auxiliary loss along with classification loss i.e. $L + L_{ss}$ where $L_{ss} \in \{L_e, L_{rot}\}$. 
\vspace{-3pt}
\paragraph{Step 2: Fine-tuning with Manifold Mixup:} Fine-tune the above model with Manifold Mixup loss $L_{mm}$ for a few more epochs. 

After obtaining the backbone, a cosine classifier is learned over it to adapt to few-shot tasks. \textit{S2M2$_R$} and \textit{S2M2$_E$} are two variants of our proposed approach which uses $L_{rot}$ and $L_{e}$ as auxiliary loss in Step 1 respectively.

\begin{algorithm}[t]
\SetAlgoLined
\Begin{
\footnotesize{
    \textbf{Input}: $\{\textbf{x},y\} \in \mathcal{D}_b; \alpha; \{\textbf{x}',y'\} \in \mathcal{D}_{val}$\\
    \textbf{Output}: Backbone model $f_{\theta}$\\ 
    \Comment{Feature extractor backbone  $f_{\theta}$ training}\\
    Initialize $f_{\theta}$\\
    \For{$epochs \in \{1, 2, ..., 400\}$}
    {
        Training data of size B - $(X(i), Y(i))$. \\
        $L(\theta, X(i), Y(i)) =   L_{class} + L_{ss}$\\
        $\theta \rightarrow \theta - \eta *\nabla L(\theta, X(i), Y(i)) $
    }
    $val\_acc\_prev = 0.0$\\
    $val\_acc\_list = [$ $]$\\
    \Comment{Fine-tuning $f_{\theta}$ with Manifold Mixup}\\
    \While{$val\_acc > val\_acc\_prev$}
    {
        Training data of size B - $(X(i), Y(i))$. \\
        $L(\theta, X(i), Y(i)) =   L_{mm} + 0.5(L_{class} + L_{ss})$ \\
        $\theta \rightarrow \theta - \eta * \nabla L(\theta, X(i), Y(i)) $ \\
        $val\_acc = Accuracy_{x,y \in D_{val}}( W_n(f_{\theta}(x)) , y)$ \\
        Append $val\_acc$ to $val\_acc\_list$ \\ 
        Update $val\_acc\_prev$ with $val\_acc$
        
    }
    \textbf{return} fine-tuned backbone $f_{\theta}$.
    }
 }
 \caption{\footnotesize{\textit{S2M2} feature backbone training}}
\label{feature_adv_train_algo}
\end{algorithm}





\begin{table*}[!]
\centering
\scalebox{0.75}{
\begin{tabular}{|c|c|c|c|c|c|c|c|c|}
\hline
\textbf{Method} & \multicolumn{2}{c|}{\textbf{\textit{mini}-ImageNet}} & \multicolumn{2}{c|}{\textbf{\textit{tiered}-ImageNet}} & \multicolumn{2}{c|}{\textbf{CUB}}& \multicolumn{2}{c|}{\textbf{CIFAR-FS}}\\
 & 1-Shot & 5-Shot & 1-Shot & 5-Shot & 1-Shot & 5-Shot & 1-Shot & 5-Shot\\

\hline\hline
MAML \cite{finn2017model} & 54.69 $\pm$ 0.89 & 66.62 $\pm$ 0.83 & 51.67 $\pm$ 1.81 & 70.30 $\pm$ 0.08 & 71.29 $\pm$ 0.95 &  80.33 $\pm$ 0.70 & 58.9 $\pm$ 1.9 & 71.5 $\pm$ 1.0\\
ProtoNet \cite{snell2017prototypical} & 54.16 $\pm$ 0.82  &  73.68$\pm$0.65 & 53.31 $\pm$ 0.89 & 72.69 $\pm$ 0.74 & 71.88$\pm$0.91 &   87.42 $\pm$ 0.48 & 55.5 $\pm$ 0.7 & 72.0 $\pm$ 0.6 \\
RelationNet \cite{relation2017Sung}&  52.19 $\pm$ 0.83 & 70.20 $\pm$ 0.66 & 54.48 $\pm$ 0.93   & 71.32 $\pm$ 0.78 &  68.65 $\pm$ 0.91 &  81.12 $\pm$ 0.63 & 55.0 $\pm$ 1.0 & 69.3 $\pm$ 0.8 \\
\hline
LEO \cite{leo2019}  & 61.76 $\pm$ 0.08 & 77.59 $\pm$ 0.12 & 66.33 $\pm$ 0.05 & 81.44 $\pm$ 0.09 & 68.22 $\pm$ 0.22$^*$ & 78.27 $\pm$ 0.16$^*$ & -& - \\
DCO \cite{dco2019}& 62.64 $\pm$ 0.61 & 78.63 $\pm$ 0.46 & 65.99 $\pm$ 0.72 & 81.56 $\pm$ 0.53 & - & - & 72.0 $\pm$ 0.7 & 84.2 $\pm$ 0.5 \\
\hline
Baseline++ & 57.53 $\pm$ 0.10 & 72.99 $\pm$ 0.43 & 60.98 $\pm$ 0.21 & 75.93 $\pm$ 0.17 &  70.4 $\pm$ 0.81 & 82.92 $\pm$ 0.78 & 67.50 $\pm$ 0.64 & 80.08 $\pm$ 0.32\\
Manifold Mixup & 57.16 $\pm$ 0.17 & 75.89 $\pm$ 0.13 & 68.19 $\pm$ 0.23 & 84.61 $\pm$ 0.16 &  73.47 $\pm$ 0.89 & 85.42 $\pm$ 0.53 & 69.20 $\pm$ 0.2 & 83.42 $\pm$ 0.15\\
Rotation & 63.9 $\pm$ 0.18 & 81.03 $\pm$ 0.11 & 73.04 $\pm$ 0.22  & 87.89 $\pm$ 0.14 & 77.61 $\pm$ 0.86 & 89.32 $\pm$ 0.46 & 70.66 $\pm$ 0.2 & 84.15 $\pm$ 0.14 \\

\textit{S2M2$_R$} &  \textbf{64.93 $\pm$ 0.18} & \textbf{83.18 $\pm$ 0.11} & \textbf{73.71 $\pm$ 0.22}  & \textbf{88.59 $\pm$ 0.14} & \textbf{80.68 $\pm$ 0.81} & \textbf{90.85 $\pm$ 0.44} & \textbf{74.81 $\pm$ 0.19} & \textbf{87.47 $\pm$ 0.13} \\
\hline
\end{tabular}}
\captionof{table}{\footnotesize{Comparison with prior/current state of the art methods on \textit{mini}-ImageNet, \textit{tiered}-ImageNet, CUB and CIFAR-FS dataset. The accuracy with $^*$ denotes our implementation of LEO using their publicly released code}}
\label{table:sota-comparison}
\end{table*}

\begin{table*}[!h]
\centering
\scalebox{0.83}{
\begin{tabular}{|c|c|c|c|c|c|c|c|}
\hline
\textbf{Dataset} & \textbf{Method} & \multicolumn{2}{c|}{\textbf{ResNet-18}} & \multicolumn{2}{c|}{\textbf{ResNet-34}} & \multicolumn{2}{c|}{\textbf{WRN-28-10}} \\
 & & 1-Shot & 5-Shot & 1-Shot & 5-Shot & 1-Shot & 5-Shot \\

\hline\hline
\multirow{7}{*}{\textit{mini}-ImageNet}  & Baseline++ & 53.56 $\pm$ 0.32 & 74.02                                            $\pm$ 0.13 & 54.41 $\pm$ 0.21 & 74.14                                     $\pm$ 0.19 & 57.53 $\pm$ 0.10 & 72.99 $\pm$ 0.43\\
                                & Mixup ($\alpha=1$) & 56.12 $\pm$ 0.17 & 73.42 $\pm$ 0.13 & 56.19 $\pm$ 0.17 & 73.05 $\pm$ 0.12 & 59.65 $\pm$ 0.34& 77.52 $\pm$ 0.52 \\
                                & Manifold Mixup & 55.77 $\pm$ 0.23 & 71.15 $\pm$ 0.12  & 55.40 $\pm$ 0.37 & 70.0 $\pm$ 0.11 & 57.16 $\pm$ 0.17 & 75.89 $\pm$ 0.13 \\
                                & Rotation & 58.96 $\pm$ 0.24 & 76.63 $\pm$ 0.12 & 61.13 $\pm$ 0.2 & 77.05 $\pm$ 0.35 & 63.9 $\pm$ 0.18 & 81.03 $\pm$ 0.11\\
                                & Exemplar & 56.39 $\pm$ 0.17 & 76.33 $\pm$ 0.14 & 56.87 $\pm$ 0.17 & 76.90 $\pm$ 0.17 & 62.2 $\pm$ 0.45 & 78.8 $\pm$ 0.15 \\
                        & \textit{S2M2$_E$}  & 56.80 $\pm$ 0.2 & 76.54 $\pm$ 0.14 & 56.92 $\pm$ 0.18 & 76.97 $\pm$ 0.18 & 62.33 $\pm$ 0.25 & 79.35 $\pm$ 0.16\\
                                & \textit{S2M2$_R$} & \textbf{64.06 $\pm$ 0.18} & \textbf{80.58 $\pm$ 0.12} & \textbf{63.74 $\pm$ 0.18} & \textbf{79.45 $\pm$ 0.12} & \textbf{64.93 $\pm$ 0.18} & \textbf{83.18 $\pm$ 0.11}\\

\hline
\multirow{7}{*}{CUB}            & Baseline++ & 67.68 $\pm$ 0.23 & 82.26 $\pm$ 0.15                                 & 68.09 $\pm$ 0.23 & 83.16 $\pm$ 0.3 & 70.4 $\pm$                                  0.81 & 82.92 $\pm$ 0.78 \\
                                & Mixup ($\alpha=1$) & 68.61 $\pm$ 0.64 & 81.29 $\pm$ 0.54 & 67.02 $\pm$ 0.85 & 84.05 $\pm$ 0.5 & 68.15 $\pm$ 0.11 & 85.30 $\pm$ 0.43 \\
                                & Manifold Mixup & 70.57 $\pm$ 0.71 & 84.15 $\pm$ 0.54 & 72.51 $\pm$ 0.94 &  85.23 $\pm$ 0.51 & 73.47 $\pm$ 0.89 & 85.42 $\pm$ 0.53 \\
                                & Rotation & 72.4 $\pm$ 0.34 & 84.83 $\pm$ 0.32 & 72.74 $\pm$ 0.46 & 84.76 $\pm$ 0.62 & 77.61 $\pm$ 0.86 & 89.32 $\pm$ 0.46 \\
                                & Exemplar& 68.12 $\pm$ 0.87 & 81.87 $\pm$ 0.59 & 69.93 $\pm$ 0.37 & 84.25 $\pm$ 0.56 & 71.58 $\pm$ 0.32 & 84.63 $\pm$ 0.57 \\
                        & \textit{S2M2$_E$} & \textbf{71.81 $\pm$ 0.43} & \textbf{86.22 $\pm$ 0.53} & 72.67 $\pm$ 0.27 & 84.86 $\pm$ 0.13 & 74.89 $\pm$ 0.36 & 87.48 $\pm$ 0.49\\
                                & \textit{S2M2$_R$}  & 71.43 $\pm$ 0.28 & 85.55 $\pm$ 0.52 & \textbf{72.92 $\pm$ 0.83} & \textbf{86.55 $\pm$ 0.51} & \textbf{80.68 $\pm$ 0.81} & \textbf{90.85 $\pm$ 0.44}\\
\hline
\multirow{7}{*}{CIFAR-FS}       & Baseline++ & 59.67 $\pm$ 0.90 & 71.40 $\pm$ 0.69                                 & 60.39 $\pm$ 0.28 & 72.85 $\pm$ 0.65 & 67.5 $\pm$                                 0.64 & 80.08 $\pm$ 0.32 \\
                                & Mixup ($\alpha=1$) & 56.60 $\pm$ 0.11 & 71.49 $\pm$ 0.35 & 57.60 $\pm$ 0.24 & 71.97 $\pm$ 0.14 & 69.29 $\pm$ 0.22 & 82.44 $\pm$ 0.27 \\
                                & Manifold Mixup & 60.58 $\pm$ 0.31 & 74.46 $\pm$ 0.13 & 58.88 $\pm$ 0.21 & 73.46 $\pm$ 0.14 & 69.20 $\pm$ 0.2 & 83.42 $\pm$ 0.15\\
                                & Rotation & 59.53 $\pm$ 0.28 & 72.94 $\pm$ 0.19 & 59.32 $\pm$ 0.13 &  73.26 $\pm$ 0.15 & 70.66 $\pm$ 0.2 & 84.15 $\pm$ 0.14 \\
                                & Exemplar  &59.69 $\pm$ 0.19 & 73.30 $\pm$ 0.17 & 61.59 $\pm$ 0.31 & 74.17 $\pm$ 0.37 & 70.05 $\pm$ 0.17 & 84.01 $\pm$ 0.22 \\
                        & \textit{S2M2$_E$}  & 61.95 $\pm$ 0.11 & 75.09 $\pm$ 0.16 & 62.48 $\pm$ 0.21 & 73.88 $\pm$ 0.30 & 72.63 $\pm$ 0.16 & 86.12 $\pm$ 0.26 \\
                        & \textit{S2M2$_R$}  & \textbf{63.66$\pm$ 0.17} & \textbf{76.07$\pm$ 0.19} & \textbf{62.77$\pm$ 0.23} & \textbf{75.75$\pm$ 0.13} & \textbf{74.81 $\pm$ 0.19} & \textbf{87.47 $\pm$ 0.13}\\
\hline
\end{tabular}}
\captionof{table}{\footnotesize{Results on \textit{mini}-ImageNet, CUB and CIFAR-FS dataset over different network architecture.}}
\label{table:main-results}
\end{table*}

\section{Experiments and Results}
\label{experiments}
In this section, we present our results of few-shot classification task on different datasets and model architectures. We first describe the datasets, evaluation criteria and implementation details\footnote {\href{https://github.com/nupurkmr9/S2M2\_fewshot}{https://github.com/nupurkmr9/S2M2\_fewshot}}.
\vspace{-8pt}
\paragraph{Datasets:} 
We perform experiments on four standard datasets for few-shot image classification benchmark, \textit{mini}-ImageNet \cite{vinyals2016matching}, \textit{tiered}-ImageNet \cite{tieredImagenet}, CUB \cite{WahCUB_200_2011} and CIFAR-FS \cite{cifar-fs2018Bertinetto}. \textbf{\textit{mini}-ImageNet} consists of $100$ classes from the ImageNet \cite{imagenet2014Russakovsky} which are split randomly into $64$ base, $16$ validation and $20$ novel classes. Each class has $600$ samples of size $84\times84$. \textbf{{\textit{tiered}-ImageNet}}  consists of $608$ classes randomly picked from ImageNet \cite{imagenet2014Russakovsky} which are split randomly into $351$ base, $97$ validation and $160$ novel classes. In total, there are $779,165$ images of size $84\times84$. \textbf{CUB} contains $200$ classes with total $11,788$ images of size $84\times84$. The base, validation and novel split is $100$, $50$ and $50$ classes. \textbf{CIFAR-FS} is created by randomly splitting $100$ classes of CIFAR-$100$ \cite{cifar100dataset} into $64$ base, $16$ validation and $20$ novel classes. The images are of size $32\times32$.
\vspace{-8pt}
\paragraph{Evaluation Criteria:} 
We evaluate experiments on $5$-way $1$-shot and $5$-way $5$-shot \cite{vinyals2016matching} classification setting i.e using $1$ and $5$ labeled instances of each of the $5$ classes as training data and $Q$ instances each from the same classes as testing data. For {\textit{tiered}-ImageNet}, \textit{mini}-ImageNet and CIFAR-FS we report the average classification accuracy over $10000$ tasks where $Q=599$ for $1$-Shot and $Q=595$ for $5$-Shot tasks respectively. For CUB we report average classification accuracy with $Q=15$ over $600$ tasks. We compare our approach \textit{S2M2$_R$} against the current state-of-the-art methods, LEO \cite{leo2019} and DCO \cite{dco2019} in Section \ref{results}.

\subsection{Implementation Details}
We perform experiments on three different model architecture: ResNet-18, ResNet-34 \cite{resnet2015He} and WRN-28-10 \cite{wrn2016Zagoruyko} which is a Wide Residual Network of $28$ layers and width factor $10$. For \textit{tiered}-ImageNet we only perform experiments with WRN-28-10 architecture. Average pooling is applied at the last block of each architecture for getting feature vectors. ResNet-18 and ResNet-34 models have $512$ dimensional output feature vector and WRN-28-10 has $640$ dimensional feature vector. For training ResNet-18 and ResNet-34 architectures, we use Adam \cite{kingma2014adam} optimizer for \textit{mini}-ImageNet and CUB whereas SGD optimizer for CIFAR-FS. For WRN-28-10 training, we use Adam optimizer for all datasets.

\begin{figure*}[t]
\centering
\scalebox{.8}{
    \includegraphics[width=2.15in,height=2.7in,keepaspectratio]{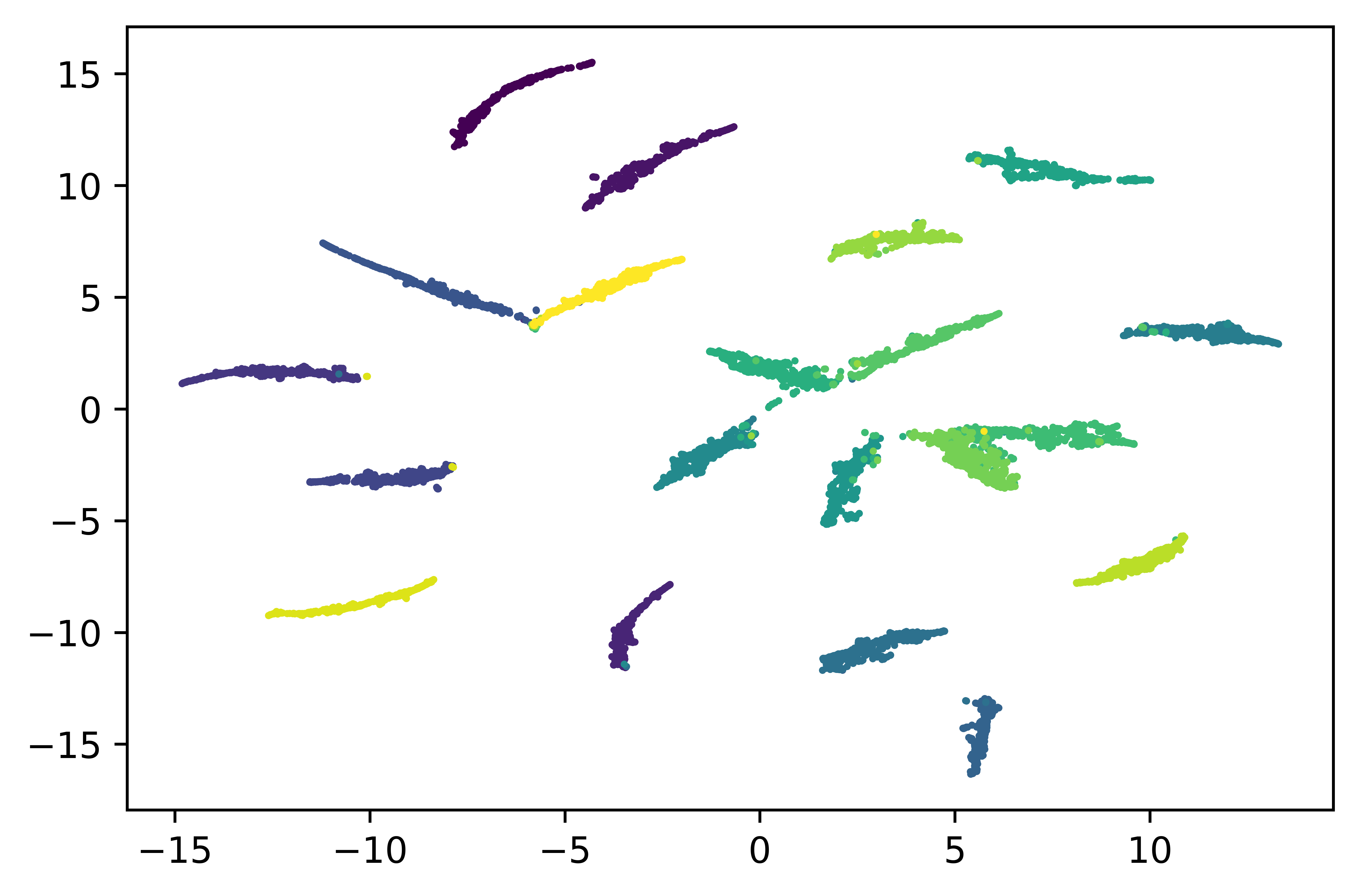} 
    \includegraphics[width=2.15in,height=2.7in,keepaspectratio]{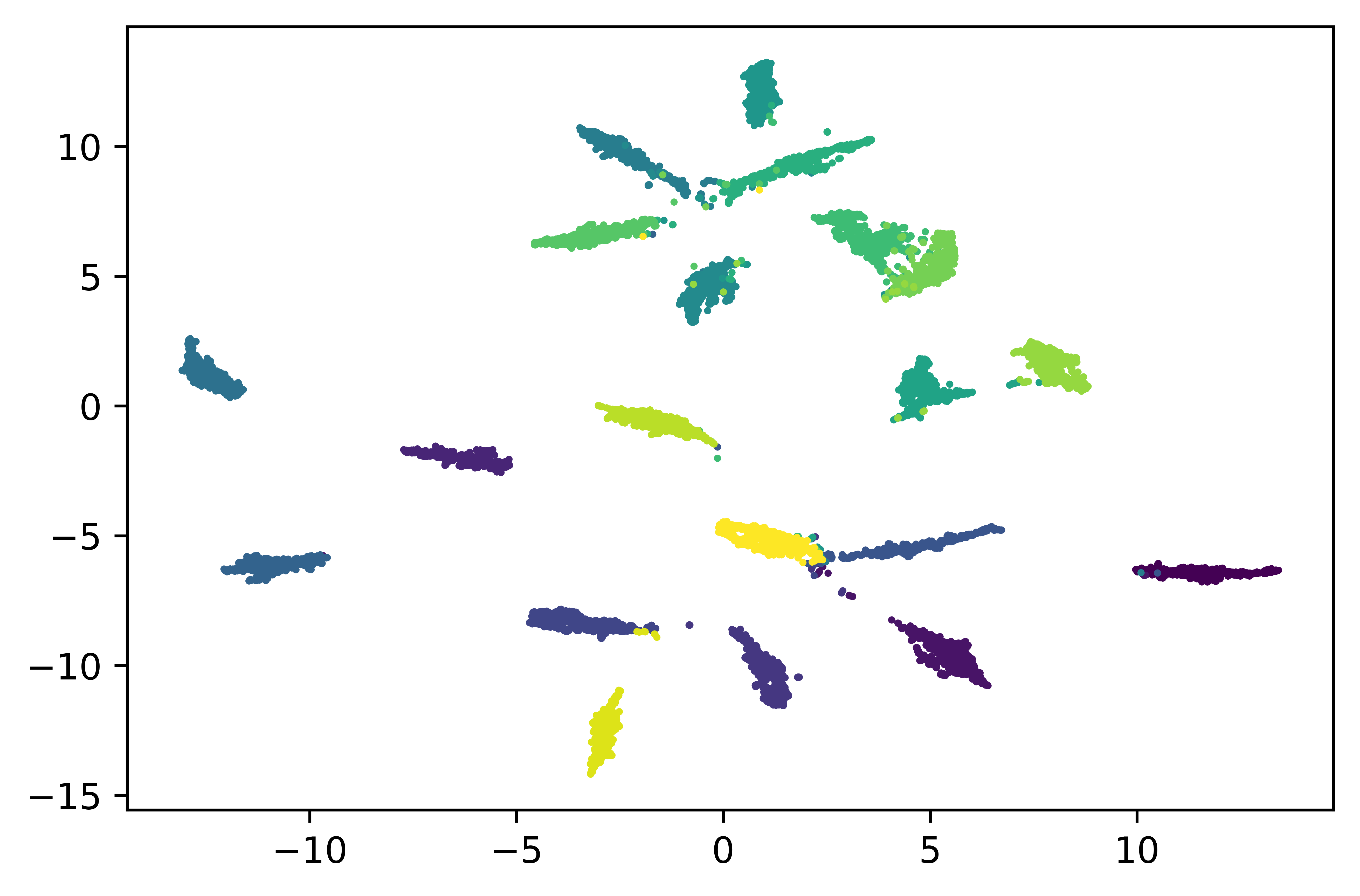} 
    \includegraphics[width=2.15in,height=2.7in,keepaspectratio]{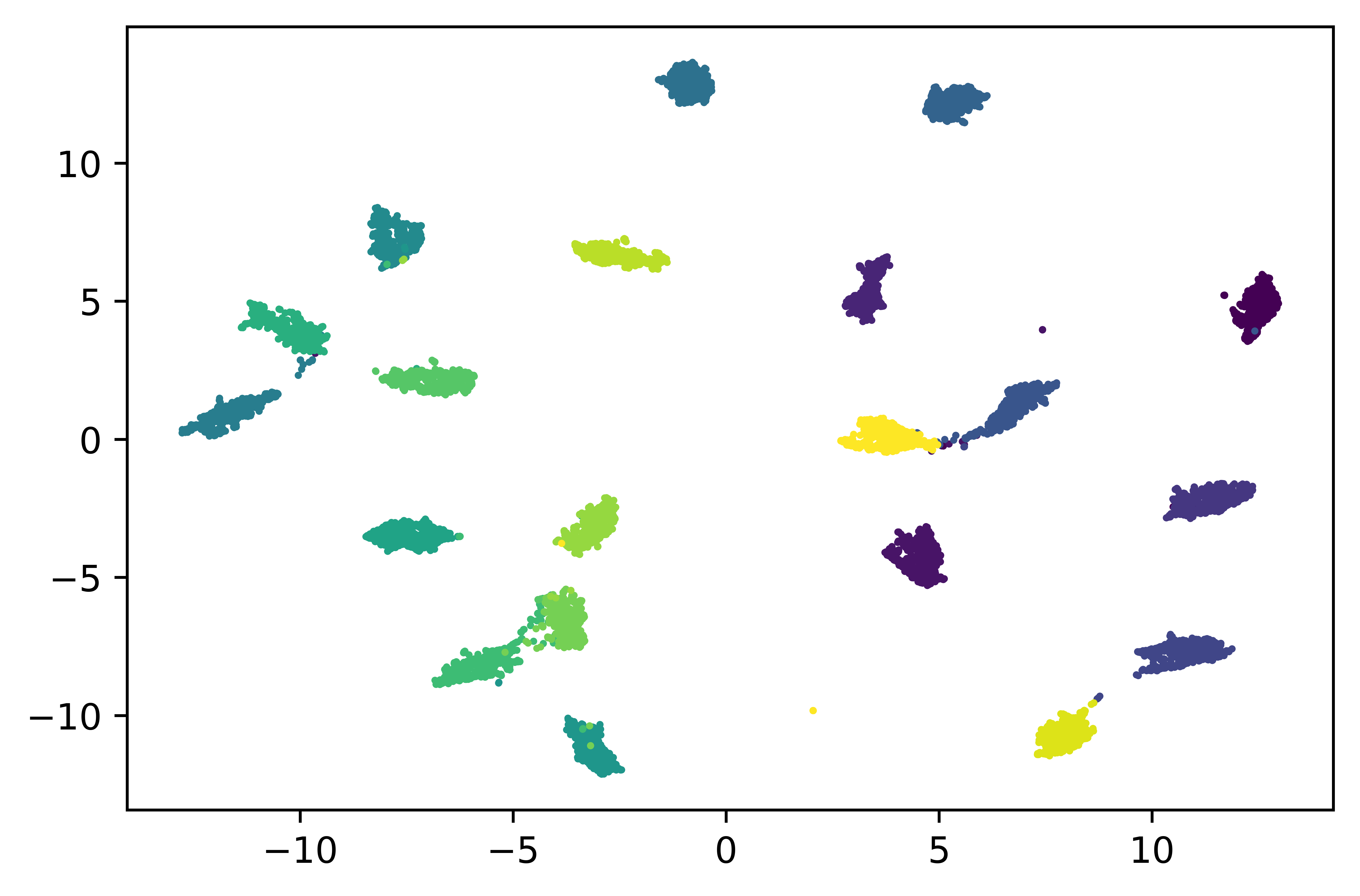}
}
    \captionof{figure}{\footnotesize{UMAP (2-dim) \cite{umaplearn} plot of feature vectors of images from novel classes of \textit{mini}-ImageNet using Baseline++, Rotation, \textit{S2M2$_R$} (left to right).}}
    \label{2-dplot}
\end{figure*}

\vspace{-3pt}
\subsection{Performance Evaluation over Few-shot Tasks}
\vspace{-5pt}
In this subsection, we report the result of few shot learning over our proposed methodology and its variants.
\vspace{-4pt}
\subsubsection{Using Manifold Mixup Regularization}
\vspace{-4pt}
All experiments using Manifold Mixup \cite{verma2019manifold} randomly sample a hidden layer (including input layer) at each step to apply mixup as described in equation \ref{eq:eqn_mixup} for the mini-batch with mixup coefficient ($\lambda$) sampled from a $\beta(\alpha , \alpha)$ distribution with $\alpha$ = $2$.
We compare the performance of Manifold Mixup \cite{verma2019manifold} with Baseline++ \cite{chen2019closerfewshot} and Mixup \cite{zhang2018mixup}. The results are shown in table \ref{table:main-results}. We can see that the boost in few-shot accuracy from the two aforementioned mixup strategies is significant when model architecture is deep (WRN-28-10). For shallower backbones (ResNet-18 and ResNet-34), the results are not conclusive.

\subsubsection{Using Self-supervision as Auxiliary Loss}
We evaluate the contribution of rotation prediction \cite{Spyros2018rotate} and exemplar training \cite{exemplar2014} as an auxiliary task during backbone training for few-shot tasks. Backbone model is trained with both classification loss and auxiliary loss as explained in section \ref{selfsupervision}. For exemplar training, we use random cropping, random horizontal/vertical flip and image jitter randomization \cite{s4l2019} to produce 4 different positive variants of each image in the mini-batch. Since exemplar training is computationally expensive, we fine-tune the baseline++ model for $50$ epochs using both exemplar and classification loss.

The comparison of above techniques with Baseline++ is shown in table
\ref{table:main-results}. As we see, by selecting rotation and exemplar as an auxiliary loss there is a significant improvement from Baseline++ ($~7-8\%$) in most cases. Also, the improvement is more prominent for deeper backbones like WRN-28-10.

\vspace{-3pt}
\subsubsection{Our Approach: \textit{S2M2}}\label{results}
We first train the backbone model using self-supervision (exemplar or rotation) as auxiliary loss and then fine-tune it with Manifold Mixup as explained in section \ref{ourapproach}.
The results are shown in table \ref{table:main-results}. We compare our approach with current state-of-the-art \cite{leo2019, dco2019} and other existing few-shot methods \cite{snell2017prototypical, relation2017Sung} in Table \ref{table:sota-comparison}. As we can observe from table, our approach \textit{S2M2$_R$} beats the most recent state-of-the-art results , LEO \cite{leo2019} and DCO \cite{dco2019}, by a significant margin on all four datasets. We find that using only rotation prediction as an auxiliary task during backbone training also outperforms the existing state-of-the-art methods on all datasets except CIFAR-FS.
{\centering
\scalebox{0.67}{
\begin{tabular}{@{}|c|l|l|l|l|l|l|l|l|@{}}
\toprule
\multirow{2}{*}{Method}& \multicolumn{2}{c|}{5-way} & \multicolumn{2}{c|}{10-way}    & \multicolumn{2}{c|}{15-way}     & \multicolumn{2}{c|}{20-way}    \\
                        & 1-shot        & 5-shot &    1-shot        & 5-shot    & 1-shot         & 5-shot         & 1-shot        & 5-shot         \\ \midrule
Baseline++           & 57.53         & 72.99    & 40.43         & 56.89          & 31.96          & 48.2           & 26.92         & 42.8           \\
LEO \cite{leo2019}        & 61.76         & 77.59            & 45.26         & 64.36          & 36.74          & 56.26          & 31.42         & 50.48          \\
DCO \cite{dco2019}           & 62.64         & 78.63              & 44.83         & 64.49          & 36.88          & 57.04          & 31.5          & 51.25          \\ \midrule
\begin{tabular}[c]{@{}c@{}}Manifold-\\ Mixup\end{tabular}   & 57.16         & 75.89        & 42.46         & 62.48          & 34.32          & 54.9           & 29.24         & 48.74          \\
Rotation         & 63.9         & 81.03        & 47.77         & 67.2           & 38.4           & 59.59          & 33.21         & 54.16          \\
\textit{S2M2$_R$}     & \textbf{64.93} & \textbf{83.18}             & \textbf{50.4} & \textbf{70.93} & \textbf{41.65} & \textbf{63.32} & \textbf{36.5} & \textbf{58.36} \\ \bottomrule
\end{tabular}
}
\captionof{table}{\footnotesize{Mean few-shot accuracy on \textit{mini}-ImageNet as $N$ increases in $N$-way $K$-shot classification.}}
\label{table:N_Way}}

\section{Discussion and Ablation Studies}
To understand the significance of learned feature representation for few-shot tasks, we perform various experiments and analyze the findings in this section.  We choose \textit{mini}-ImageNet as the primary dataset with WRN-28-10 backbone for the following experiments.
\paragraph{Effect of varying $N$ in $N$-way classification:}
For extensive evaluation, we test our proposed methodology in complex few-shot settings. We vary $N$ in $N$-way $K$-shot evaluation criteria from $5$ to $10$, $15$ and $20$. The corresponding results are reported in table \ref{table:N_Way}. We observe that our approach \textit{S2M2$_R$} outperforms other techniques by a significant margin. The improvement becomes more pronounced for $N>5$. Figure \ref{2-dplot} shows the 2-dimensional UMAP \cite{umaplearn} plot of feature vectors of novel classes obtained from different methods. It shows that our approach has more segregated clusters with less variance. This supports our hypothesis that using both self supervision and Manifold Mixup regularization helps in learning feature representations with well separated margin between novel classes. 
\paragraph{Cross-domain few-shot learning:}
We believe that in practical scenarios, there may be a significant domain-shift between the base classes and novel classes. Therefore, to further highlight the significance of selecting the right manifold for feature space, we evaluate the few-shot classification performance over cross-domain dataset : \textbf{\textit{mini}-ImageNet $\implies$ CUB} (coarse-grained to fine-grained distribution) using Baseline++, Manifold Mixup \cite{verma2019manifold}, Rotation \cite{s4l2019} and \textit{S2M2$_R$}.
We train the feature backbone with the base classes of \textit{mini}-ImageNet  and evaluate its performance over the novel classes of CUB (to highlight the domain-shift). We report the corresponding results in table \ref{table:cross-domain}.

\begin{table}[t]
\centering
\scalebox{0.95}{
\begin{tabular}{|c|c|c|c|c|c|c|c|}
\hline
\textbf{Method} & \multicolumn{2}{c|}{\textbf{\textit{mini}-ImageNet} $\implies$ \textbf{CUB}} \\
 & 1-Shot & 5-Shot \\
 \hline \hline
DCO \cite{dco2019}  & 44.79 $\pm$ 0.75 & 64.98 $\pm$ 0.68 \\
\hline
Baseline++ & 40.44 $\pm$ 0.75 & 56.64 $\pm$ 0.72\\
Manifold Mixup  & 46.21 $\pm$ 0.77 & 66.03 $\pm$ 0.71 \\
Rotation & \textbf{48.42 $\pm$ 0.84} & 68.40 $\pm$ 0.75 \\
\textit{S2M2$_R$}  & 48.24 $\pm$ 0.84 & \textbf{70.44 $\pm$ 0.75}\\

\hline
\end{tabular}}
\captionof{table}{\footnotesize{Comparison in cross-domain dataset scenario.}}
\label{table:cross-domain}
\end{table}

\begin{table}[t]
\centering
\scalebox{0.95}{
\begin{tabular}{|c|c|c|c|c|c|c|c|}
\hline
\textbf{Method} & \multicolumn{2}{c|}{\textbf{Base + Validation}}\\
 & 1-Shot & 5-Shot \\
 \hline \hline
LEO \cite{leo2019} & 61.76 $\pm$ 0.08 & 77.59 $\pm$ 0.12 \\
DCO \cite{dco2019}  & 64.09 $\pm$ 0.62 &  80.00 $\pm$ 0.45 \\
\hline
Baseline++ & 61.10 $\pm$ 0.19 & 75.23 $\pm$ 0.12\\
Manifold Mixup & 61.10 $\pm$ 0.27 & 77.69 $\pm$ 0.21\\
Rotation & {65.98 $\pm$ 0.36} & {81.67 $\pm$ 0.08} \\ 
\textit{S2M2$_R$} & \textbf{67.13 $\pm$ 0.13} & \textbf{83.6 $\pm$ 0.34}\\
\hline
\end{tabular}
}
\captionof{table}{\footnotesize{Effect of using the union of base and validation class for training the backbone $f_{\theta}$.}}
\label{table:validation}
\end{table}
\paragraph{Generalization performance of supervised learning over base classes:}
The results in table \ref{table:main-results} and \ref{table:N_Way} empirically support the hypothesis that our approach learns a feature manifold that generalizes to novel classes and also results in improved performance on few-shot tasks. This generalization of the learned feature representation should also hold for base classes. To investigate this, we evaluate the performance of backbone model over the validation set of the ImageNet dataset and the recently proposed ImageNetV2 dataset\cite{recht2019imagenet}. ImageNetV2 was proposed to test the generalizability of the ImageNet trained models and consists of images having slightly different data distribution from the ImageNet. We further test the performance of backbone model over some common visual perturbations and adversarial attack. We randomly choose $3$ of the $15$ different perturbation techniques - pixelation, brightness, contrast , with $5$ varying intensity values , as mentioned in the paper \cite{hendrycks2019robustness}. For adversarial attack, we use the FGSM \cite{goodfellow2014explaining} with $\epsilon = 1.0/255.0 $. All the evaluation is over the $64$ classes of \textit{mini}-ImageNet used for training the backbone model. The results are shown in table \ref{table:robustness}. It can be seen that \textit{S2M2$_R$} 
has the best generalization performance for the base classes also.

\paragraph{Effect of using the union of base and validation classes:}
We test the performance of few-shot tasks after merging the validation classes into base classes. In table \ref{table:validation}, we see a considerable improvement over the other approaches using the same extended data, supporting the generalizability claim of the proposed method.

\begin{figure}[]
\centering
\scalebox{1.}{
    \includegraphics[width=2.0in,height=1.4in]{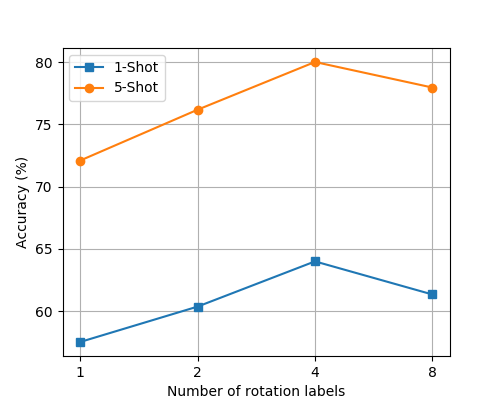} 
}
    \captionof{figure}{\footnotesize{Effect of increasing the number of self-supervised (degrees of rotation) labels.}}
    \label{fig:more-labels}
\end{figure}

\paragraph{Different levels of self-supervision:}
We conduct a separate experiment to evaluate the performance of the model by varying the difficulty of self-supervision task; specifically the number of angles to predict in rotation task. We change the number of rotated versions of each image to 1 ($0^{\circ}$), 2 ($0^{\circ}$, $180^{\circ}$), 4 ($0^{\circ}$,$90^{\circ}$,$180^{\circ}$,$270^{\circ}$) and 8 ($0^{\circ}$,$45^{\circ}$,$90^{\circ}$,$135^{\circ}$,$180^{\circ}$,$225^{\circ}$,$270^{\circ}$,$315^{\circ}$) and record the performance over the novel tasks for each of the corresponding $4$ variants. Figure \ref{fig:more-labels} shows that the performance improves with increasing the number of rotation variants till $4$, after which the performance starts to decline.

\begin{table}[t]
\centering
\scalebox{0.9}{
\begin{tabular}{|c|c|c|c|c|c|c|}
\hline
Methods & I & I2 & P & C & B & Adv \\ \hline
Baseline++                                                          & 80.75          & 81.47          & 70.54                                                      & 47.11                                                    & 74.36          & 19.75                                                       \\ \hline
Rotation                                                            & 82.21          & 83.91          & 71.9                                                       & 50.84                                                    & 76.26          & 20.5                                                        \\ \hline
\begin{tabular}[c]{@{}c@{}}Manifold\\ Mixup\end{tabular}                                                      & 83.75          & 87.19          & 75.22                                                      & 57.57                                                    & 78.54          & \textbf{44.97}                                              \\ \hline
\textit{S2M2$_R$} & \textbf{85.28} & \textbf{88.41} & \textbf{75.66}                                             & \textbf{60.0}                                            & \textbf{79.77} & 28.0                                                        \\ \hline
\end{tabular}
}
\captionof{table}{\footnotesize{Validation set top-1 accuracy of different approaches over base classes and it's perturbed variants (I:ImageNet; I2:ImageNetv2; P:Pixelation noise; C: Contrast noise; B: Brightness; Adv: Aversarial noise)}}
\label{table:robustness}
\end{table}

\label{discussion}
\section{Conclusion}
We observe that learning feature representation with relevant regularization and self-supervision techniques lead to consistent improvement of few-shot learning tasks on a diverse set of image classification datasets. 
Notably, we demonstrate that feature representation learning using both self-supervision and classification loss and then applying Manifold Mixup over it, outperforms prior state-of-the-art approaches in few-shot learning. We do extensive experiments to analyze the effect of architecture and efficacy of learned feature representations in few-shot setting. This work opens up a pathway to further explore the techniques in self-supervision and generalization techniques to improve computer vision tasks specifically in low-data regime. Finally, our findings highlight the merits of learning a robust representation that helps in improving the performance of few-shot tasks.
{\small
\bibliographystyle{ieee}
\bibliography{egbib}
}

\twocolumn[\appendixhead]

\section{Ablation Studies}
In this section, we perform additional experiments to study the efficacy of our approach \textit{S2M2$_R$}.
\subsection{Effect of varying $N$ in $N$-way classification on CIFAR-FS}
We vary $N$ in $N$-way $K$-shot evaluation criteria from $5$ to $10$, $15$ and $20$ for CIFAR-FS dataset. The corresponding results are reported in table \ref{table:N_Waycifar}. We observe that our approach \textit{S2M2$_R$} outperforms other techniques by a significant margin. The improvement becomes more pronounced for $N>5$. Figure \ref{cifar2-dplot} shows the 2-dimensional UMAP \cite{umaplearn} plot of feature vectors of novel classes obtained from different methods. We obtain similar results for CIFAR-FS as that in the case of \textit{mini}-ImageNet. We show that our approach has more segregated clusters with less variance. This supports our hypothesis that using both self-supervision and Manifold Mixup regularization helps in learning feature representations with well separated margin between novel classes.

\subsection{Visualizing important regions in images responsible for classification}
We visualize the relevant pixels responsible for classifying a particular image to the correct class. We define the relevance of the pixels as the top-1 percentile of the pixels sorted by the magnitude of the gradient with respect to the correct class of the image. For this experiment, we use models trained using the Baseline++ \cite{chen2019closerfewshot} and \textit{S2M2$_R$} methods to visualize the relevant pixels. In figure \ref{attr}, we show the relevant pixels of the image \textit{highlighted in white} for visualization. Qualitatively speaking, we observe that relevant pixels for model trained using \textit{S2M2$_R$} tends to focus more on the object belonging to the specified class and not in the background.

\begin{minipage}{\textwidth}
\centering
    \includegraphics[width=1.5in,height=1.5in]{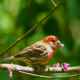} 
    \hspace{0.05cm}
    \includegraphics[width=1.5in,height=1.5in]{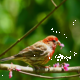}
    \hspace{0.05cm}
    \includegraphics[width=1.5in,height=1.5in]{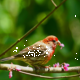}
    
    \includegraphics[width=1.5in,height=1.5in]{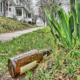}
    \hspace{0.05cm}
    \includegraphics[width=1.5in,height=1.5in]{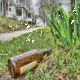}
    \hspace{0.05cm}
    \includegraphics[width=1.5in,height=1.5in]{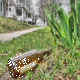}
    
    \includegraphics[width=1.5in,height=1.5in]{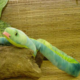} 
    \hspace{0.05cm}
    \includegraphics[width=1.5in,height=1.5in]{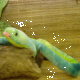}
    \hspace{0.05cm}
    \includegraphics[width=1.5in,height=1.5in]{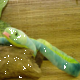}

    \includegraphics[width=1.5in,height=1.5in]{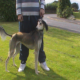}
    \hspace{0.05cm}
    \includegraphics[width=1.5in,height=1.5in]{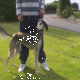}
    \hspace{0.05cm}
    \includegraphics[width=1.5in,height=1.5in]{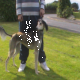}

    \captionof{figure}{\footnotesize{Each row visualizes the relevant pixels for classification with respect to a trained model for a image sampled from the base class of \textit{mini}-ImageNet ( house finch, beer bottle, green mamba, Saluki). Images in each row are arranged in the order with labels as original image, relevant pixels by Baseline++ model and relevant pixels by \textit{S2M2$_R$} model \textit{(from left to right)} respectively. The relevant pixels is defined as the Top-1 percentile of pixels responsible for classification \textit{(pixels marked in white color)}}.
    }
    \label{attr}
\end{minipage}

\end{document}